\documentclass[10pt,twocolumn,letterpaper]{article}

\usepackage{cvpr}
\usepackage{times}
\usepackage{epsfig}
\usepackage{graphicx}
\usepackage{amsmath,bm}
\usepackage{amssymb}

\usepackage{booktabs,caption}
\usepackage[flushleft]{threeparttable}

\usepackage[pagebackref=true,breaklinks=true,colorlinks,bookmarks=false]{hyperref}

\cvprfinalcopy 

\def\cvprPaperID{****} 

\ifcvprfinal\fi
\begin{document}

\title{Detailed Human Shape Estimation from a Single Image\linebreak by Hierarchical Mesh Deformation}

\author{
Hao Zhu$^{1,2}$ \quad Xinxin Zuo$^{2,3}$ \quad Sen Wang$^{3,2}$ \quad Xun Cao$^{1}$ \quad Ruigang Yang$^{2,4,5}$\\
$^{1}$Nanjing University, Nanjing, China \quad $^{2}$University of Kentucky, Lexington, KY, USA\\
$^{3}$Northwestern Polytechnical University, Xi'an, China \quad $^{4}$Baidu Inc., Beijing, China\\
$^{5}$National Engineering Laboratory of Deep Learning and Technology and Application, China\\
{\tt\small zhuhao\_nju@163.com \quad xinxin.zuo@uky.edu \quad wangsen1312@gmail.com}\\
{\tt\small caoxun@nju.edu.cn \quad yangruigang@baidu.com}
}

\maketitle

\begin{abstract}
	This paper presents a novel framework to recover \emph{detailed} human body shapes from a single image. It is a challenging task due to factors such as variations in human shapes, body poses, and viewpoints. Prior methods typically attempt to recover the human body shape using a parametric based template that lacks the surface details. As such the resulting body shape appears to be without clothing. In this paper, we propose a novel learning-based framework that combines the robustness of parametric model with the flexibility of free-form 3D deformation. We use the deep neural networks to refine the 3D shape in a Hierarchical Mesh Deformation (HMD) framework, utilizing the constraints from body joints, silhouettes, and per-pixel shading information. We are able to restore detailed human body shapes beyond skinned models. Experiments demonstrate that our method has outperformed previous state-of-the-art approaches, achieving better accuracy in terms of both 2D IoU number and 3D metric distance.  The code is available in \url{https://github.com/zhuhao-nju/hmd.git}.

	
\end{abstract}

\vspace{-10pt}
\section{Introduction}

Recovering 3D human shape from a single image is a challenging problem and has drawn much attention in recent years. A large number of approaches~\cite{ICCV2009Guan, ECCV2016Bogo, 3DV2016Dibra, BMVC2017Tan, CVPR2017Lassner, NIPS2017Tung, CVPR2018Pavlakos, CVPR2018Kanazawa, 3DV2018Omran} have been proposed in which the human body shapes get reconstructed by predicting the parameters of a statistical body shape model, such as SMPL~\cite{TOG2015Loper} and SCAPE~\cite{TOG2005Anguelov}. The parametric shape is of low-fidelity, and unable to capture clothing details. 
Another collection of methods\cite{ECCV2018Varol, BMVC2018Venkat} estimate volumetric human shape directly from the image using CNN, while the resulting volumetric representation is fairly coarse and does not contain shape details.

\begin{figure}[t]
	\begin{center}
		\includegraphics[width=1.0\linewidth]{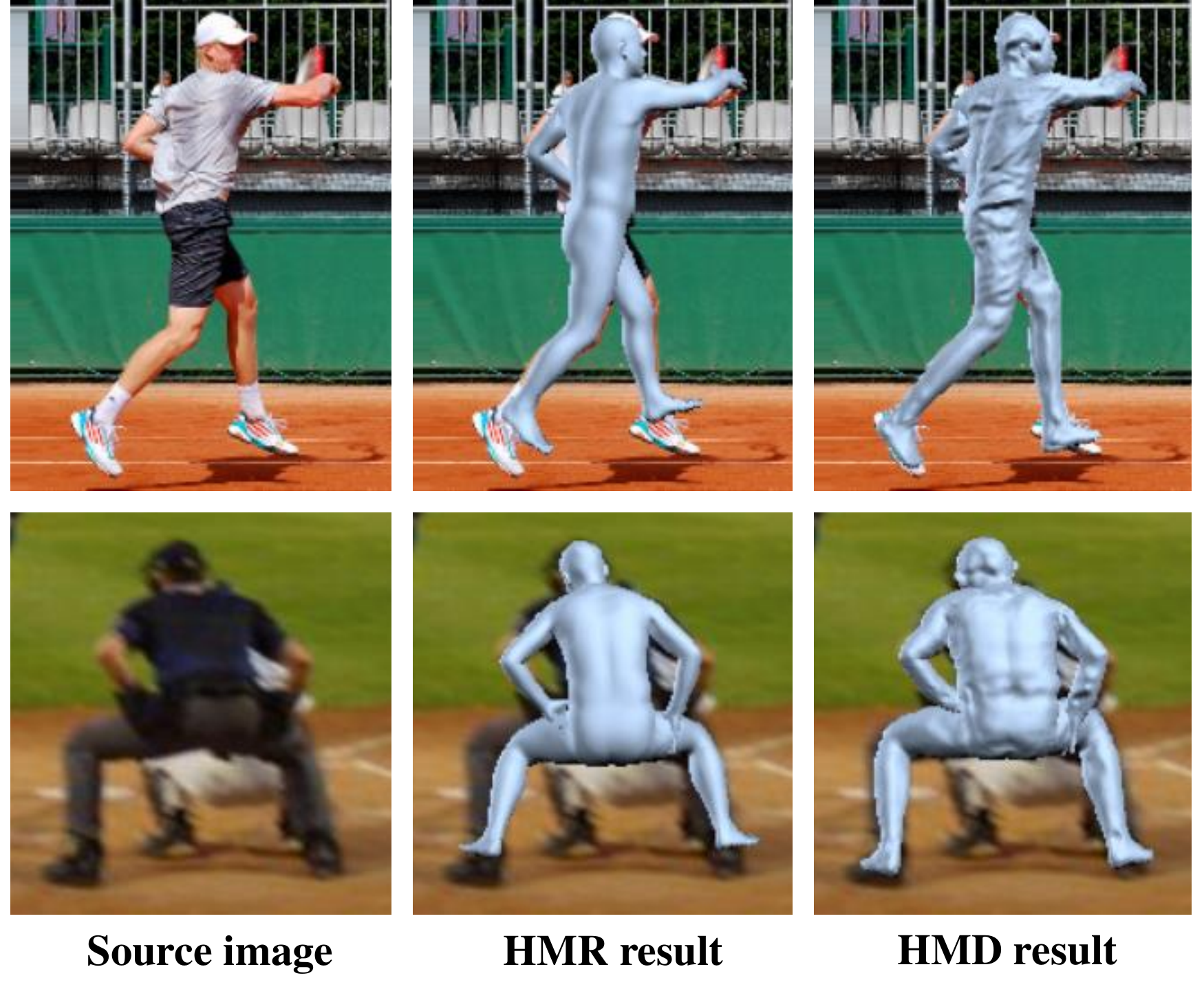}
		\vspace{-15pt}
	\end{center}
	\caption{Our method (HMD) takes a single 2D image of a person as input and predicts detailed human body shape. As compared with the current state-of-the-art method (HMR~\cite{CVPR2018Kanazawa}), we have got the recovered body shapes with surface details that better fit to the input image.}
	\label{fig:tile}
	\vspace{-5pt}
\end{figure}

The limited performance of previous methods is caused by the large variations of the human shape and pose. Parametric or volumetric shapes are not expressive enough to represent the inherent complexity. 




In this paper, we propose a novel framework to reconstruct \emph{detailed} human shape from a single image.  The key here is to combine the robustness of parametric model with the flexibility of free-form deformation. In short, we build on top of current SMPL model to obtain an initial parametric mesh model and perform non-rigid 3D deformation on the mesh to refine the surface shape.
We design a coarse-to-fine refinement scheme in which a deep neural network is used in each stage to estimate the 3D mesh vertex movement by minimizing its 2D projection error in the image space. 
We feed window-cropped images instead of full image to the network, which leads to more accurate and robust prediction of deformation.
In addition, we integrate a photometric term to allow high-frequency details to be recovered. 
These techniques combined lead to a method that significantly improves, both visually and quantitatively, the recovered human shape from a single image as shown in Figure \ref{fig:tile}.

The contributions of this paper include:
\begin{itemize}
	\vspace{-0.1in}
	\item 
	We develop a novel \textit{project - predict - deform} strategy to predict the deformation of the 3D mesh model by using 2D features. 
	
	\vspace{-0.1in}
	\item 
	We carefully design a hierarchical update structure, incorporating body joints, silhouettes, and photometric-stereo to improve shape accuracy without losing the robustness. 
	
	
	\vspace{-0.1in}
	\item 
	We are the first to use a single image to recover detailed human shape beyond parameters human model. As demonstrated throughout our experiments, the additional free deformation of the initial parametric model leads to quantitatively more accurate shapes with good generalization capabilities to images in the wild.   
	
\end{itemize}

\begin{figure*}[t]
	\begin{center}
		\includegraphics[width=0.95\linewidth]{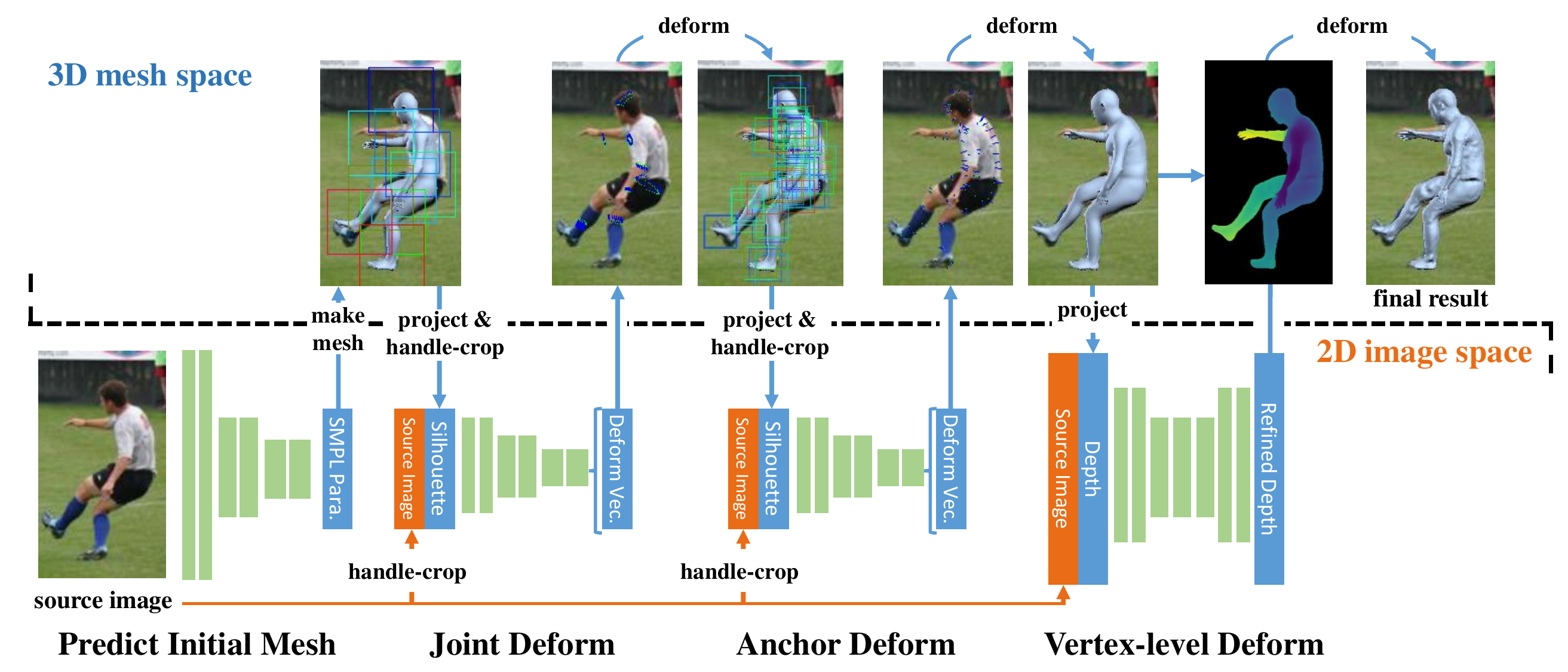}
		\vspace{-10pt}
	\end{center}
	\caption{The flow of our method goes from the bottom left to the top right.  The mesh deformation architecture consists of three levels: joint, anchor and per-vertex.  In each level, the 3D mesh is projected to 2D space and sent together with the source image to the prediction network. The 3D mesh gets deformed by the predicted results to produce refined human body shapes.}
	\vspace{-10pt}
	\label{fig:pipeline}
\end{figure*}

\section{Related Work}
Previous approaches can be divided into two categories based on the way the human body is represented: parametric methods and non-parametric methods. 

As for parametric methods, they rely on a pre-trained generative human model, such as the SCAPE~\cite{TOG2005Anguelov} or SMPL~\cite{TOG2015Loper} models. The goal is to predict the parameters of the generative model. The SCAPE model has been adopted by Guan \etal~\cite{ICCV2009Guan} to recover the human shape and poses from the monocular image as provided with some manually clicked key points and the constraint of smooth shading.  Instead of relying on manual intervention, Dibra \etal \cite{3DV2016Dibra} have trained a convolutional neural network to predict SCAPE parameters from a single silhouette. Similar to the SCAPE model, Hasler \etal \cite{CVPR2010Hasler} have proposed a multilinear model of human pose and body shape that is generated by factorizing the measurements into the pose and shape dependent components. The SMPL model~\cite{TOG2015Loper} has recently drawn much attention due to its flexibility and efficiency. For example, Bogo \etal~\cite{ECCV2016Bogo} have presented an automatic approach called SMPLify which fits the SMPL model by minimizing an objective function that penalizes the error between the projected model joints and detected 2D joints obtained from a CNN-based method together with some priors over the pose and shape. Building upon this SMPLify method, Lassner \etal~\cite{CVPR2017Lassner} have formed an initial dataset of 3D body fitting with rich annotations consisting of 91 keypoints and 31 segments. Using this dataset, they have shown improved performance on part segmentation, pose estimation and 3D fitting. Tan \etal~\cite{BMVC2017Tan} proposed an indirect learning procedure by first training a decoder to predict body silhouettes from SMPL parameters and then using pairs of real images and ground truth silhouettes to train a full encoder-decoder network to predict SMPL parameters at the information bottleneck. Pavlakos \etal~\cite{CVPR2018Pavlakos} separated the SMPL parameters prediction network into two sub-networks. Taking the 2D image as input, the first network was designed to predict the silhouette and 2D joints, from which the shape and pose parameters were estimated respectively. The latter network combined the shape and 2D joints to predict the final mesh. Kanazawa \etal~\cite{CVPR2018Kanazawa} proposed an end-to-end framework to recover the human body shape and pose in the form of SMPL model using only 2D joints annotations with an adversarial loss to effectively constrain the pose. Instead of using joints or silhouettes. Omran \etal~\cite{3DV2018Omran} believed that a reliable bottom-up semantic body part segmentation was more effective for shape and pose prediction. Therefore, they predicted a part segmentation from the input image in the first stage and took this segmentation to predict SMPL parameterization of the body mesh.



Non-parametric methods directly predict the shape representation from the image. 
Some researchers have used depth maps as a more general and direct representation for shapes. For example, Varol \etal~\cite{CVPR2017Varol} trained a convolutional neural network by building up a synthetic dataset of rendered SMPL models to predict the human shape in the form of depth image and body part segmentation.  G{\"u}ler \etal~\cite{CVPR2017Guler, CVPR2018Guler} have treated the shape prediction problem as a correspondence regression problem, which would produce a dense 2D-to-3D surface correspondence field for the human body. Another way of representing 3D shapes is to embed the 3D mesh into a volumetric space~\cite{ECCV2018Varol,BMVC2018Venkat}. For example, Varol \etal~\cite{ECCV2018Varol} restored volumetric body shape directly from a single image. 
Their method focuses more on robust body measurements rather than shape details or poses.


While significant progress has been made in this very difficult problem, the resulting human shape is still lacking in accuracy and details, visually they all look like undressed.

\section{Hierarchical Deformation Framework}
We present our hierarchical deformation framework to recover \emph{detailed} human body shapes by refining a template model in a coarse-to-fine manner. As shown in Figure~\ref{fig:pipeline}, there are four stages in our framework: First, an initial SMPL mesh is estimated from the source image. Starting from this, the next three stages serve as refinement phases which predict the deformation of the mesh so as to produce a detailed human shape. We have used the HMR method~\cite{CVPR2018Kanazawa} to predict the initial human mesh model, which has demonstrated state-of-the-art performance on human shape recovery from a single image. However, like other human shape recovery methods~\cite{CVPR2018Pavlakos, 3DV2018Omran, ECCV2016Bogo} that utilize the SMPL model, the HMR method predicts the shape and pose parameters to generate a skinned mesh model with limited flexibility to closely fit the input image or express surface details. For example, the HMR often predicts deflected joint position of limbs when the human pose is unusual.  Therefore, we have designed our framework to refine both the shape and the pose. 


The refining stages are arranged hierarchically from coarse to fine. We define three levels of key points on the mesh, referred to as \emph{handles} in this paper. We will describe exactly how we define these handles in the next section. In each level, we design a deep neural network to refine the 3D mesh geometry using these handles as control points. We train the three refinement networks separately and successively to predict the residual deformation based on its previous phase.

To realize the overall refinement procedure, a challenging problem is how to deform the 3D human mesh from handles in 2D space using deep neural networks. We address this using Laplacian mesh deformation. In detail, the motion vector for each handle is predicted from the network driven by the joints and silhouettes of the 2D image. Then the human mesh will get deformed with the Laplacian deformation approach given the movements of the handles while maintaining the local geometry as much as possible. The deforming strategy has been used in multi-view shape reconstruction problem~\cite{CVPR2018Alldieck, TCSVT2017Zhu, liao2009modeling, ICCV2001Plankers, zhu2017role, ECCV2016Rhodin, 3DV2016Robertini}, while we are the first to predict the deformation from a single image with the deep neural network.

\subsection{Handle Definitions}
In this section, we will describe the handles that we have used in each level. They could be predefined in the template model thanks to the uniform topology of SMPL mesh model.



\begin{figure}[t]
	\begin{center}
		\includegraphics[width=1.0\linewidth]{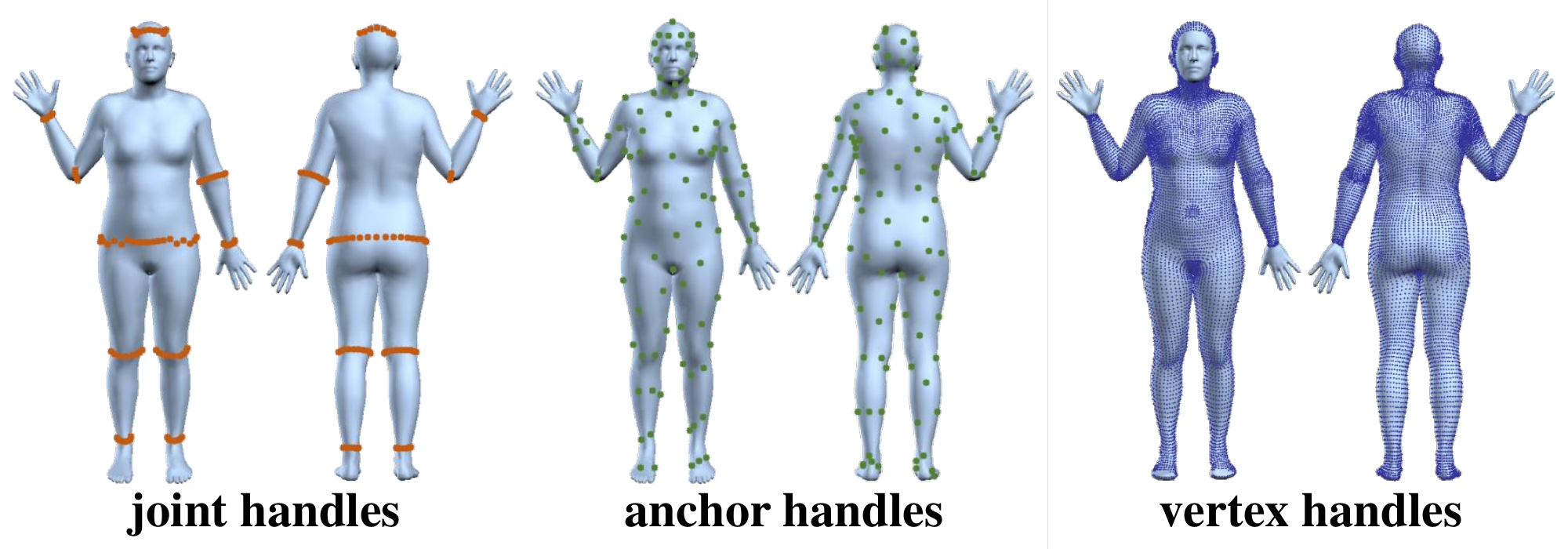}
		\vspace{-0.4in}
	\end{center}
	\caption{The handles definition in different levels for mesh deformation.  
	}
	\vspace{-0.2in}
	\label{fig:handles}
\end{figure}

\textbf{Joint handles.}  We select 10 joints as the control points -- head, waist, left/right shoulders, left/right elbows, left/right knees, and left/right ankles. The vertices around the joints under the T-pose SMPL mesh are selected as handles, as shown in Figure~\ref{fig:handles}. We take the geometric center of each set of handles as the position of its corresponding body joint. The motion of each joint handle is represented as a 2D vector, which refers to the vector from the joint position of projected mesh to ground truth joint position on the image plane.

\textbf{Anchor handles.}  We select 200 vertices on the human template mesh under T-pose as anchor handles. To select the anchors evenly over the template, we firstly build a vector set $C = \{v_1, v_2, ......, v_n\}$ with $v_i$ concatenated by the position and surface normal of the vertex $i$ and $n$ is the number of SMPL model vertices. Then K-means is applied to cluster set $C$ into 200 classes. Finally, we set the closest vertex to the center of each cluster as the anchor handles. Besides, we have removed the vertices in the face, fingers, toes from the T-pose SMPL model to prevent the interference of high-frequency shape. To be noticed that, for each anchor, it is only allowed to move along the surface normal direction, so we just need to predict a single value indicating the movement of the anchor point along the normal direction.


\textbf{Vertex-level handles.} The vertices in the SMPL mesh are too sparse to apply pixel-level deform, so we subdivide each face of the mesh into 4 faces.  The subdivision increases the number of vertices of the template mesh to 27554, and all these vertices are regarded as handles. 

\subsection{Joint and Anchor Prediction}
\label{sec:deformation}

\textbf{Network.} Both joint and anchor prediction networks use the VGG~\cite{arXiv2014VGG} structure which consists of a feature extractor and a regressor. The network takes the mesh-projected silhouette and source image as input, which are cropped into patches as centered with our predefined handles. Specifically, for a $224\times224$ input image, the image is cropped into patches with the size of $64\times64$ for joint prediction, and $32\times32$ for anchor prediction. Comparing to the full image or silhouette as input, the handle cropped input allows the network to focus on the region of interest. 
We will demonstrate the effectiveness of the cropped input in Section \ref{sec:stage_exp}.  

\textbf{Loss.}  The output of the joint net is 2D vector representing the joint motion in the image plane.  L2 loss is exploited to train the joint net with the loss function formulated as:

\vspace{-0.12in}
\begin{equation}
	L_{joint} = ||\bm{p} - \bm{\widehat{p}}||_2
	\label{equ:loss_func}
\end{equation}
\vspace{-0.12in}

where $\bm{p}$ is the predicted motion vector from the network and $\widehat{\bm{p}}$ is the displacement vector from the mesh-projected joint position to its corresponding ground truth joint. Both vectors are 2-dimensional.


As for the anchor net, our immediate goal is to minimize the area of the mismatched part between the projected silhouette and the ground truth silhouette.  However, as it is hard to compute the derivatives of area size with respect to the motion vector for back-propagation in the training process, we transform the mismatched area size to the length of the projected line segment along vertex normal direction which falls into the mismatched region.  The searching radius of the segment line is $0.1m$.  The length of this line segment is regarded as the ground truth anchor movement, and L2 loss is used to train the network.  This conversion makes it easy to calculate the gradient for the loss function.  If the ground truth movement of one anchor handle is zero, this handle would be disabled, which means the anchor would be regarded as common vertices in the Laplacian edit.  We compulsively set the anchors which are far from the silhouette margin to be inactive, and the overall shape would be deformed equably since the Laplacian deforming will keep the local geometry as much as possible.

Besides, instead of using the RGB image as input, the joint and anchor prediction network could also take ground truth silhouette of the human figure as input if available.  The silhouette provides more explicit information for the human figure, which prevents the network from getting confused by the cluttered background environment. We will demonstrate its effectiveness on joint and anchor deformation prediction in the experiment section. In this paper, we consider the RGB-only as input by default, and use `+Sil.' to indicate the case where the additional silhouette is used.




\subsection{Vertex-level Deformation}


To add high-frequency details to the reconstructed human models, we exploit the shading information contained in the input image. First, we project the current 3D human model into the image space, from which we will get the coarse depth map. We then train a \emph{Shading-Net} that takes the color image and current depth map as input and predicts a refined depth map with surface details. 
We have built up a relatively small dataset which contains color images, over-smoothed depth maps, and corresponding ground truth depth maps that have good surface details. More detailed explanations on this dataset could be found in Section \ref{ssec:Imp}.
We adopt a multi-stage training scheme with limited supervised data. 


Following the training scheme proposed in~\cite{CVPR2018sfs}, we firstly train a simple UNet based encoder-decoder network~\cite{Unet2015} on our captured depth dataset taking the over-smoothed depth map and its corresponding color image as input. The network is trained as supervised by the ground truth depth maps. Next, we apply this network on the real images of our human body dataset to obtain enhanced depth maps. As we only have limited supervised data, the network may not generalize well to our real images. Therefore, to finally get depth maps with great surface details consistent with the color images, we train our \emph{Shading-Net}, which is also a U-Net based network on real images. The network is trained with both the supervision loss using the depth maps output by the first U-Net and also a photometric reconstruction loss~\cite{ECCV2018DDRNet} that aims to minimize the error between the original input image and the reconstructed image. The per-pixel photometric loss $L_{photo}$ is formulated as below:

\vspace{-0.15in}
\begin{equation}
	\label{Eq:lighting}
	L_{photo} =||\rho \sum_{k=1}^{9} l_{k}H_{k}(\bm{n}) -I ||_2
	\vspace{-0.07in}
\end{equation}

where $\rho$ is the albedo computed by a traditional intrinsic decomposition method~\cite{bell2014intrinsic}. Similar to~\cite{ramamoorthi2001efficient, zuo2017detailed}, we use the second spherical harmonics (SH) for illumination representation under the Lambertian surface assumption. $H_{k}$ represents the basis of spherical harmonics. $l_1,l_2...l_9$ denote the SH coefficients, which are computed under a least square minimization as: 
\vspace{-0.1in}
\begin{equation}  
	\label{Eq:lighting2}
	\bm{l}^{*} = \mathop{\arg\min}_{\bm{l}}  ||\rho \sum_{k=1}^{9} l_{k}H_{k}(\bm{n}_{coarse}) -I ||_2^2
	\vspace{-0.1in}
\end{equation}
We use the coarse depth map rendered from the currently recovered 3D model to compute the surface normal $\bm{n}_{coarse}$. 


\subsection{Implementation Details}\label{ssec:Imp}
We use the pre-trained model in the HMR-net, then train Joint-Net, Anchor-Net, and Shading-Net successively.  We use the `Adam' optimizer to train these networks, with the learning rate as $0.0001$.  The handle weight in Laplacian edit is $10$ for joint deforming and is $1$ for anchor deforming.

To provide better training data to the Shading-Net, we have captured a small depth dataset with a Kinect V2 sensor. It consists of $2200$ depth frames with three human subjects wearing different clothes under various poses. The captured depth maps are further enhanced using traditional shading refinement techniques~\cite{or2015rgbd, zollhofer2014real} to recover small surface details, which can be taken as ground truth depth maps for supervised learning. We have magnified the shape details by $10$ times during the test time.

\section{Experiment}

\begin{figure*}[t]
	\begin{center}
		\includegraphics[width=0.935\linewidth]{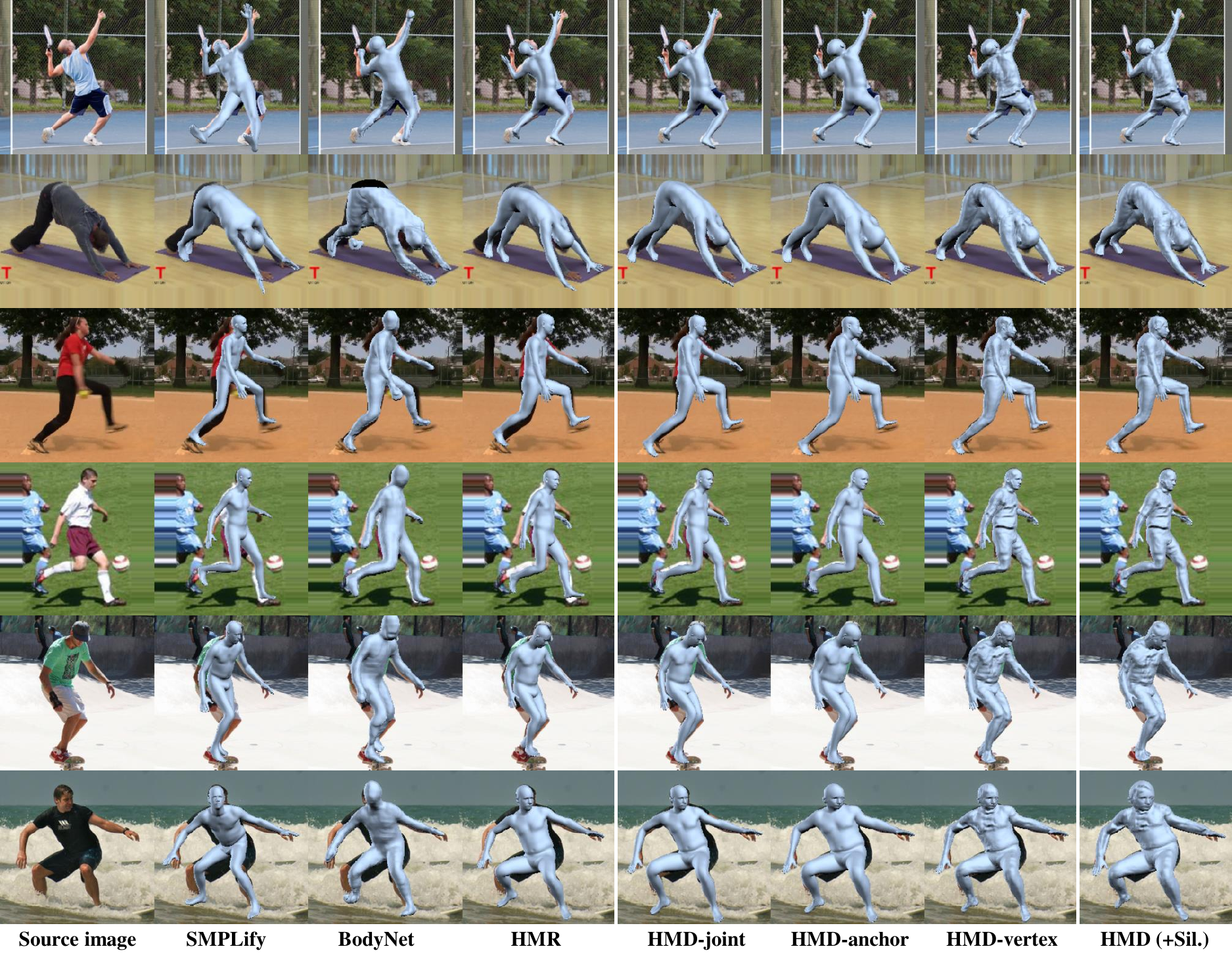}
		\vspace{-0.2in}
	\end{center}
	\caption{We compare our method with previous approaches. The results of our method in different stages are shown: joint deformed, anchor deformed and vertex deformed (final result). Comparing to other methods, our method recovers more accurate joints and the body with shape details. The human body shape fits better to the input image, especially in body limbs. The rightmost column shows we can get more accurate recovered shapes when ground truth human silhouette is enforced (labeled as \textit{HMD (+Sil.)}). Note that the images are cropped for the compact layout.}
	\vspace{-0.1in}
	\label{fig:compare}
\end{figure*}

\begin{figure*}[t]
	\begin{center}
		\includegraphics[width=0.935\linewidth]{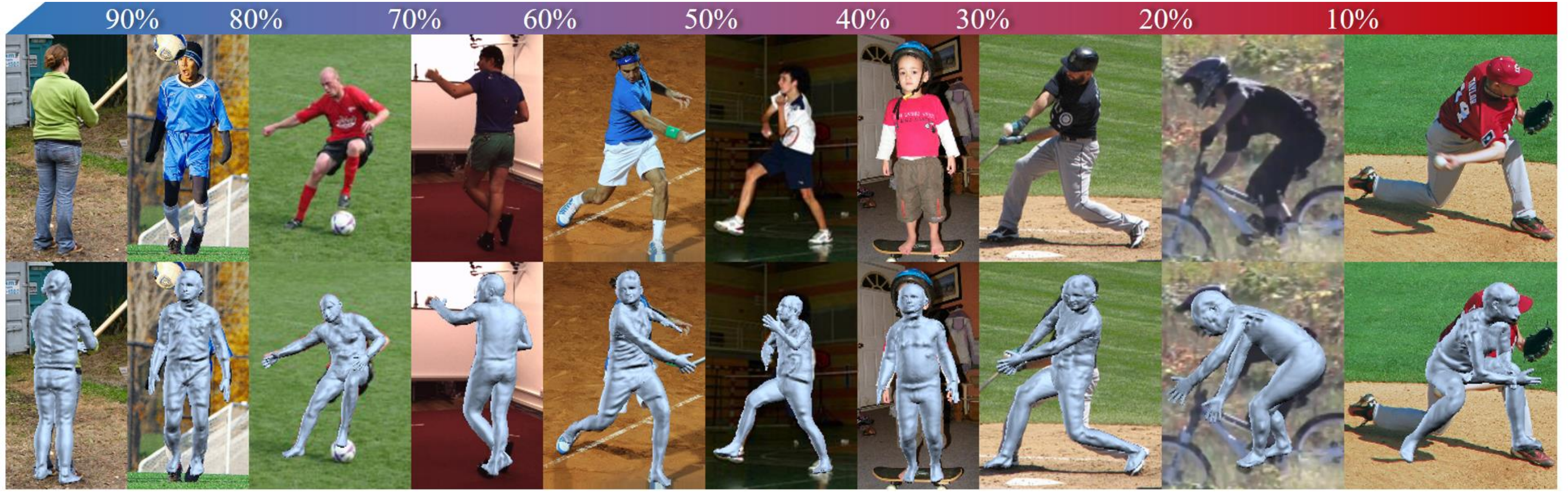}
		\vspace{-0.2in}
	\end{center}
	\caption{The results selected based on the rank of silhouette IoU.  We could see in columns of the left side, the person with a simple pose like standing yields really good fit. As we go from left to right columns, sports in the images are getting more complicated and the corresponding human shape is harder to predict. And in the right side columns, our method fails to predict humans with accessories (helmet, gloves) and under extremely twisting poses.}
	\vspace{-0.1in}
	\label{fig:perc}
\end{figure*}

\begin{figure*}[t]
	\begin{center}
		\includegraphics[width=1.0\linewidth]{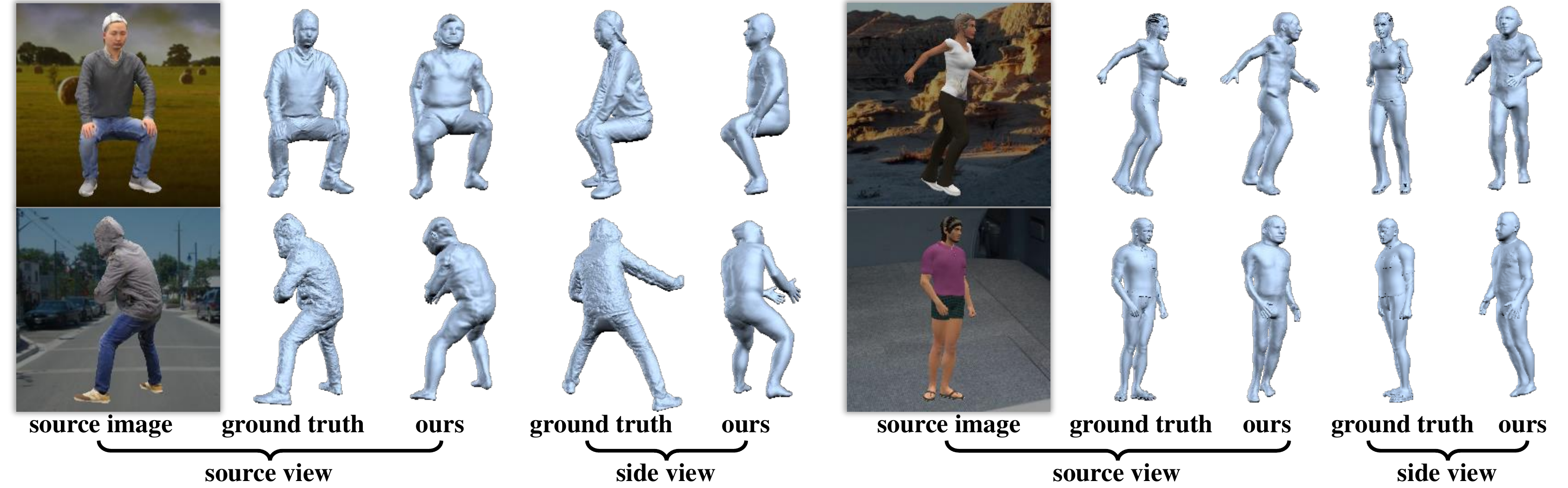}
		\vspace{-0.35in}
	\end{center}
	\caption{We show some recovered meshes and the ground truth meshes on the RECON (left) and SYN dataset (right).  The meshes are rendered in the side view by rotating the mesh $90^{\circ}$ around the vertical axis.}
	\label{fig:real_syn}
	\vspace{-0.1in}
\end{figure*}

\begin{figure*}[t]
	\begin{center}
		\includegraphics[width=1.0\linewidth]{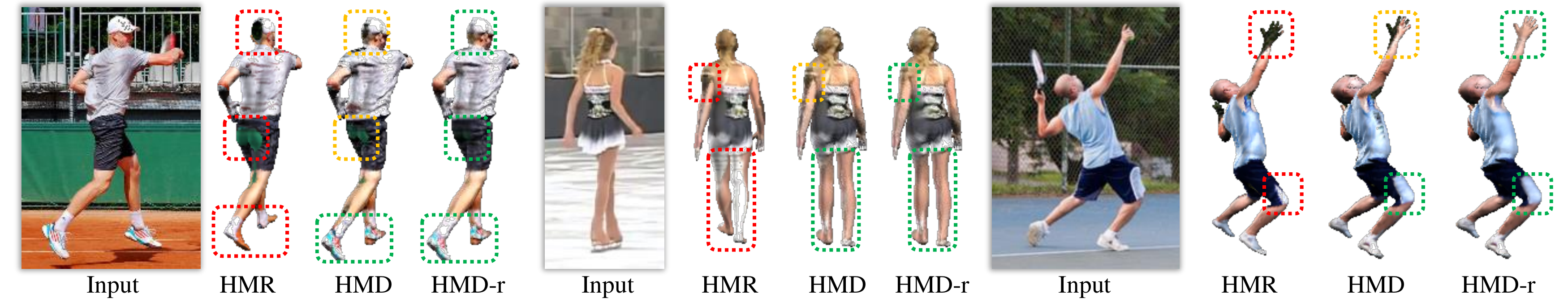}
	\end{center}
	\vspace{-0.15in}
	\caption{We show some textured results in the side view.  We directly map the texture in the image to the mesh, and render them to the novel view, as shown in HMR and HMD.  In HMD-r, we simply dilute the foreground region in the image to the background, then map the diluted image to the mesh.  The novel view is set by rotating original view with $45^\circ$ around the vertical axis crossing the mesh center.  The red dotted box means bad part, the yellow one means better but defective part, the green one means fine part. }
	\vspace{-0.15in}
	\label{fig:side_view}
\end{figure*}


\begin{table*}[]
	\begin{threeparttable}
		\caption{Quantitative Evaluation}
		\vspace{-0.1in}
		\centering
		\begin{tabular}{lcccccccc}
			\hline
			& \multicolumn{2}{c}{-----WILD dataset-----} & \multicolumn{3}{c}{ ------------RECON dataset------------} & \multicolumn{3}{c}{ --------------SYN dataset--------------} \\
			method & sil IoU  & 2D joint err  & sil IoU  & 3D err full*  & 3D err vis*  & sil IoU  & 3d err full*  & 3d err vis*  \\ \hline
			SMPLify\cite{ECCV2016Bogo}      & 66.3   & 10.19   & 73.9   & 52.84   & 51.69   & 71.0   & 62.31   & 60.67   \\
			BodyNet\cite{ECCV2018Varol}     & 68.6   & ---     & 72.5   & 43.75   & 40.05   & 70.0   & 54.41   & 46.55   \\
			HMR\cite{CVPR2018Kanazawa}      & 67.6   & 9.90    & 74.3   & 51.74   & 42.05   & 71.7   & 53.03   & 47.75   \\
			HMD - joint                    & 70.7   & 8.81    & 78.0   & 51.08   & 41.42   & 75.9   & 49.25   & 45.70   \\
			HMD - anchor                   & 76.5   & 8.82    & 85.0   & 44.60   & 39.73   & 79.6   & 47.18   & 44.62   \\
			HMD - vertex                  & ---   & ---    & ---   & 44.10   & 41.76   & ---   & 44.75   & 41.90   \\ \hline
			HMD(s) - joint                 & 73.0   & 8.31    & 79.2   & 50.49   & 40.88   & 77.7   & 48.41   & 45.16   \\
			HMD(s) - anchor                & \textbf{82.4}   & \textbf{8.22}    & \textbf{88.3}   & 43.50   & \textbf{38.63}   & \textbf{85.7}   & 44.59   & 42.68   \\
			HMD(s) - vertex               & ---   & ---    & ---   & \textbf{43.22}   & 40.98   & ---   & \textbf{41.48}   & \textbf{39.11}   \\ \hline
		\end{tabular}
		\begin{tablenotes}
			\small
			\item * `full' means the full body shape is used for error estimation, and `vis' means only the visible part with respect to the input image viewpoint is used for error estimation.
			\item The statistic unit of 3D error is millimeter; the 2D joint error is measured by pixel.  The methods beyond the cutting line use only RGB image as input, while the methods under the cutting line use `RGB + silhouette' as input.  Some statistic is blank: the joint position cannot be derived directly from the mesh produced by BodyNet;  The sil IoU and 2D joint error after vertex deformation stay the same as anchor deformed results. 
		\end{tablenotes}
		\label{tab:quantitative}
		\vspace{-0.1in}
	\end{threeparttable}
\end{table*}

\subsection{Datasets}
We have assembled three datasets for the experiment: the WILD dataset which has a large number of images with sparse 2D joints and segmentation annotated, and two other small datasets for evaluation in 3D metrics.

\textbf{WILD Dataset.}  We assemble a quite large dataset for training and testing by extracting from 5 human datasets including MPII human pose database (MPII)~\cite{CVPR2014Mykhaylo}, Common Objects in Context dataset (COCO)~\cite{ECCV2014Lin}, Human3.6M dataset (H36M)~\cite{TPAMI2014Ionescu, ICCV2011Ionescu}, Leeds Sports Pose dataset (LSP)~\cite{BMVC2010Johnson} and its extension dataset (LSPET)~\cite{CVPR2011Johnson}. As most of the images are captured in an uncontrolled setup, we call it the WILD dataset. The Unite the People (UP) dataset~\cite{CVPR2017Lassner} provides ground truth silhouettes for the images in LSP, LSPET, and MPII datasets.  As we focus on human shape recovery of the whole body, images with partial human bodies are removed based on the following two rules:

\begin{itemize}
	\vspace{-0.05in}
	\item All joints exist in the images.
	\vspace{-0.05in}
	\item All joints are inside the body silhouette.
\end{itemize}
\vspace{-0.05in}

For COCO and H36M dataset, we further filter out the data with low-quality silhouettes.
We separate the training and testing data according to the rules of each dataset.  The numbers of the data we use are listed in Table \ref{tab:datasets}.  


\begin{table}[]
	\centering
	\caption{WILD dataset components}
	\vspace{-0.12in}
	\begin{tabular}{ccccccc}
		\hline
		data source   & LSP & LSPET & MPII & COCO & H36M  \\ \hline
		train num & 987 & 5376  & 8035 & 4004 & 5747  \\
		test num  & 703 & 0     & 1996 & 606  & 1320  \\ \hline
	\end{tabular}
	\label{tab:datasets}
	\vspace{-0.1in}
\end{table}

The main drawback of WILD dataset is the lack of 3D ground truth shape.  Though the UP dataset provides the fitted SMPL mesh for some data, the accuracy is uncertain.  To help evaluate the 3D accuracy, we make two other small datasets with ground truth shape:

\textbf{RECON Dataset}  
We reconstruct 25 human mesh models using the traditional multi-view 3D reconstruction methods~\cite{FTCGV2015Furukawa}. 
We render each model to 6 views and the views are randomly selected from 54 candidate views, of which the azimuth ranges from 0$^\circ$ to $340^\circ$ with intervals $20^\circ$, and the elevation ranges from $-10^\circ$ to $+10^\circ$ with intervals of $10^\circ$.  We use various scene images from the Places dataset\cite{TPAMI2017Zhou} as background. 

\textbf{SYN Dataset}  
We render 300 synthetic human mesh models in PVHM dataset~\cite{CVPR2018Zhu} following their rendering setup, with the random scene images from Places dataset as background. The meshes of PVHM include the inner surface, which is a disturbance for surface accuracy estimation. To filter out the inner surface, we project the mesh to 6 orthogonal directions and remove the faces which are not seen in all 6 directions.  

For RECON dataset and SYN dataset,
the reconstructed 3D meshes are scaled so that the mean height of the human body are close to the general body height of the common adult. In this way, we could measure the 3D error in mm.

\begin{table}[]
	\caption{Ablation Experiments}
	\vspace{-0.1in}
	\centering
	\begin{tabular}{clcc}
		\hline
		num & method                 & sil IoU   & 2D joint err  \\ \hline
		1   & baseline(initial shape)      & 67.6      & 9.90          \\
		2   & joint (f)                    & 68.3      & 9.85          \\
		3   & joint (w)                    & 70.7      & 8.81          \\
		4   & anchor (f)                   & 70.1      & 9.89          \\
		5   & anchor (w)                   & 71.3      & 9.75          \\
		6   & joint (w) + anchor (w)       & 76.5      & 8.82          \\ \hline
	\end{tabular}
	\begin{tablenotes}
		\small
		\item The 2D joint error is measured by pixel.
	\end{tablenotes}
	\vspace{-0.2in}
	\label{tab:ablation}
\end{table}

\subsection{Accuracy Evaluations}  
We measure the accuracy of the recovered shape with several metrics (corresponding to the second row in Table~\ref{tab:quantitative}). 
For all test sets, we report the silhouette Intersection over Union (sil IoU), which measures the matching rate of the projected silhouette of the predicted 3D shape and the image silhouette.
For the WILD dataset, we measure the difference between the projected 2D joints of the predicted 3D shape and the annotated ground truth joints.  The joints of the mesh are extracted by computing the geometric center of the corresponding joint handle vertices.  For the RECON dataset and SYN dataset, we also report the 3D error (3D err), which is the average distance of vertices between the predicted mesh and the ground truth mesh. We find the closest vertices in the resulting mesh for each vertex in the ground truth mesh and compute the mean of their distances as the 3D error. 


\subsection{Staging Analysis}  
\vspace{-5pt}
\label{sec:stage_exp}

We show the staging results in Figure \ref{fig:compare} (right four columns) and report the quantitative evaluation of each stage in Table \ref{tab:quantitative}.  The results in different phases are named as HMD-joint, HMD-anchor, and HMD-vertex (final result).  We can see that the shape has got refined stage by stage.  In the joint deformation phase, the joint correction takes effects to correct the displacement of joints. In the anchor deformation phase, silhouette supervision plays a key role in fitting the human shape. In the vertex deformation stage, the shape details are recovered to produce a visually plausible result.

\textbf{Ablation study.}  We report the result of the ablation experiment in Table \ref{tab:ablation}, where (w) means the window-cropped input, and (f) means the full image input.  We demonstrate two following statements: (1) By comparing the performance between full image input (No. 3 and 5) and window-crop image input (No. 3 and 5) in the table, we could see that the window-crop input predicts much higher silhouette IoU and lower joint error comparing to full image input, while the model size of the window-crop network is only $41\%$ of the full image network.  
The reason why it has got better result is that the window-crop network inherently focuses on the handle as the input center, so the problem turns to predicting the local fit for each handle, which is easier to learn.
(2) By comparing the performance between the integration of `joint + anchor' deformation (No. 6) and only anchor or joint deformation (No. 3 and 5), we find that the combination achieves the best performance, and shows larger improvement than the pure anchor deformation. 


\textbf{Prediction with silhouette.} Our method takes the RGB image as input by default, while we could also take additional silhouettes as input. They can share the same framework and the difference is explained in Section \ref{sec:deformation}. We show the qualitative comparison result in the last column in Figure \ref{fig:compare} and the quantitative result in the last three rows in Table \ref{tab:quantitative}. As expected, the prediction with silhouette produces better results in all metrics.

\subsection{Comparison with Other Methods}
We compare our method with other methods with qualitative results shown in Figure \ref{fig:compare} and quantitative results in Table \ref{tab:quantitative}.  We use the trained model of BodyNet and HMR provided by the authors.  As BodyNet requires 3D shape for training, they don't use COCO and H36M datasets.   To be fair, the evaluation on the WILD datasets only uses the data from LSP, LSPET, and MPII, which are the intersection of datasets used in all estimated methods. 
Comparing to SMPL based methods (SMPLIify and HMR), our method has got the best performance in all metrics on all three datasets. As compared with BodyNet, a volumetric based prediction method, we have got comparable scores in 3D error on RECON dataset. The reason is that the BodyNet produces more conservative shapes instead of focusing on the recovery of a complete human model. In some cases, the body limbs have not got reconstructed by the BodyNet when they are not visible from the image, while we always have the complete body recovered even though some parts of limbs haven't appeared in the image. This makes it easy to have a better registration to the ground truth mesh resulting in smaller 3D error. However, their scores on SYN datasets are lower than the other two datasets, since the human subjects from the SYN dataset generally have slim body shapes in which case the BodyNet results are degraded.

\subsection{3D Error Analysis}

Figure \ref{fig:real_syn} shows our recovered 3D model on the RECON and SYN datasets together with the ground truth mesh. We show that the inherent pose and shape ambiguities cannot be resolved with the image from a single viewpoint. As we can see in Figure \ref{fig:real_syn}, the human shapes seen from the side view are quite different from the ground truth model even though they could fit closely to the input image. The estimated depth cue from a single image is sometimes ambiguous for shape recovery.  This observation explains the reason why the improvement of our method in 2D metrics is relatively larger than the improvement in 3D metrics. 

\subsection{View Synthesis}

We show some view synthesis results by mapping the texture in the image to the recovered mesh model in Figure \ref{fig:side_view}.  From the side view, we can see that our method yields better textured model as the mesh matches the image well.  By diluting the foreground region to the background (HMD-r), the texture in the margin parts are further improved.

\section{Conclusion}

In this paper, we have proposed a novel approach to reconstruct detailed human body shapes from a single image in a coarse-to-fine manner. Starting from an SMPL model based human recovery method, we introduce free-form deformations to refine the body shapes with a project-predict-deform strategy.  A hierarchical framework has been proposed for restoring more accurate and detailed human bodies under the supervision of joints, silhouettes, and shading information. 
We have performed extensive comparisons with state-of-the-art methods and demonstrated significant improvements in both quantitative and qualitative assessments.

The limitation of our work is that the pose ambiguities are not solved, and there are still large errors in predicted body meshes especially in depth direction. 

\vspace{-5pt}
\section*{Acknowledgements}
\vspace{-5pt}
This work was supported by the USDA grant 2018-67021-27416 and NSFC grant 61627804.

\clearpage

{\small
\bibliographystyle{ieee_fullname}
\bibliography{HumanShape}
}
\end{document}